\documentclass[conference]{IEEEtran}
\IEEEoverridecommandlockouts
\usepackage{cite}
\usepackage{amsmath,amssymb,amsfonts}
\usepackage{graphicx}
\usepackage{textcomp}
\usepackage{amsfonts}       
\usepackage{nicefrac}       
\usepackage{multirow}
\usepackage{mathrsfs}
\usepackage{wrapfig}
\usepackage{color}
\usepackage{algorithm}
\usepackage{algpseudocode}
\usepackage{setspace}
\usepackage{marvosym}
\usepackage{xcolor}
\usepackage{booktabs} 
\usepackage{bbm}
\def\BibTeX{{\rm B\kern-.05em{\sc i\kern-.025em b}\kern-.08em
    T\kern-.1667em\lower.7ex\hbox{E}\kern-.125emX}}

\newtheorem{thm}{Theorem}

\newcommand{\cX}{\mathcal{X}}
\newcommand{\ind}[1]{\mathbbm{1}_{\{#1\}}}

\begin{document}

\title{Reward Once, Penalize Once: Rectifying Time Series Anomaly Detection}

\author{\IEEEauthorblockN{Keval Doshi}
\IEEEauthorblockA{\textit{Department of Electrical Engineering)} \\
\textit{University of South Florida)}\\
Tampa, USA \\
kevaldoshi@usf.edu}
\and
\IEEEauthorblockN{Shatha Abudalou}
\IEEEauthorblockA{\textit{Department of Electrical Engineering)} \\
\textit{University of South Florida)}\\
Tampa, USA \\
sabudalou@usf.edu}
\and
\IEEEauthorblockN{Yasin Yilmaz}
\IEEEauthorblockA{\textit{Department of Electrical Engineering)} \\
\textit{University of South Florida)}\\
Tampa, USA \\
yasiny@usf.edu}

}

\maketitle

\begin{abstract}
While anomaly detection in time series has been an active area of research for several years, most recent approaches employ an inadequate evaluation criterion leading to an inflated F1 score. We show that a rudimentary Random Guess method can outperform state-of-the-art detectors in terms of this popular but faulty evaluation criterion. In this work, we propose a proper evaluation metric that measures the timeliness and precision of detecting sequential anomalies.
Moreover, most existing approaches are unable to capture temporal features from long sequences. Self-attention based approaches, such as transformers, have been demonstrated to be particularly efficient in capturing long-range dependencies while being computationally efficient during training and inference. We also propose an efficient transformer approach for anomaly detection in time series and extensively evaluate our proposed approach on several popular benchmark datasets.
\end{abstract}


\section{Introduction}
\label{introduction}
Time series analysis is used to perform important tasks such as predicting the future values of a variable (e.g., stock market price) and detecting anomalies in sequential data. Time series anomaly detection methods aim to identify abnormal data patterns in temporal data. For instance, in health care, ECG signals are analyzed to determine if the patient suffers from a heart disease \cite{ecg}. Similarly, in cybersecurity, the data traffic over time in a computer network is monitored to detect cyber-attacks \cite{doshi2021timely}. Time series anomaly detection methods are used in various domains by companies, e.g., Extensible Generic Anomaly Detection System (EGADS) by Yahoo \cite{yahoo} and SR/CNN developed by Microsoft \cite{microsoft}.

Time series anomalies are typically classified into two main categories, point anomalies and sequential anomalies. A point anomaly, also known as outlier, is a single data instance with unexpected value with respect to the nominal baseline. In many applications, anomalies continue for a duration with successive anomalous data instances, which is called a sequential anomaly. 
\cite{survey} presents a detailed survey about time series anomaly detection by evaluating twenty different methods based on statistical and deep learning approaches on univariate time series datasets. 

\begin{figure}
\centering
\includegraphics[width=.5\textwidth]{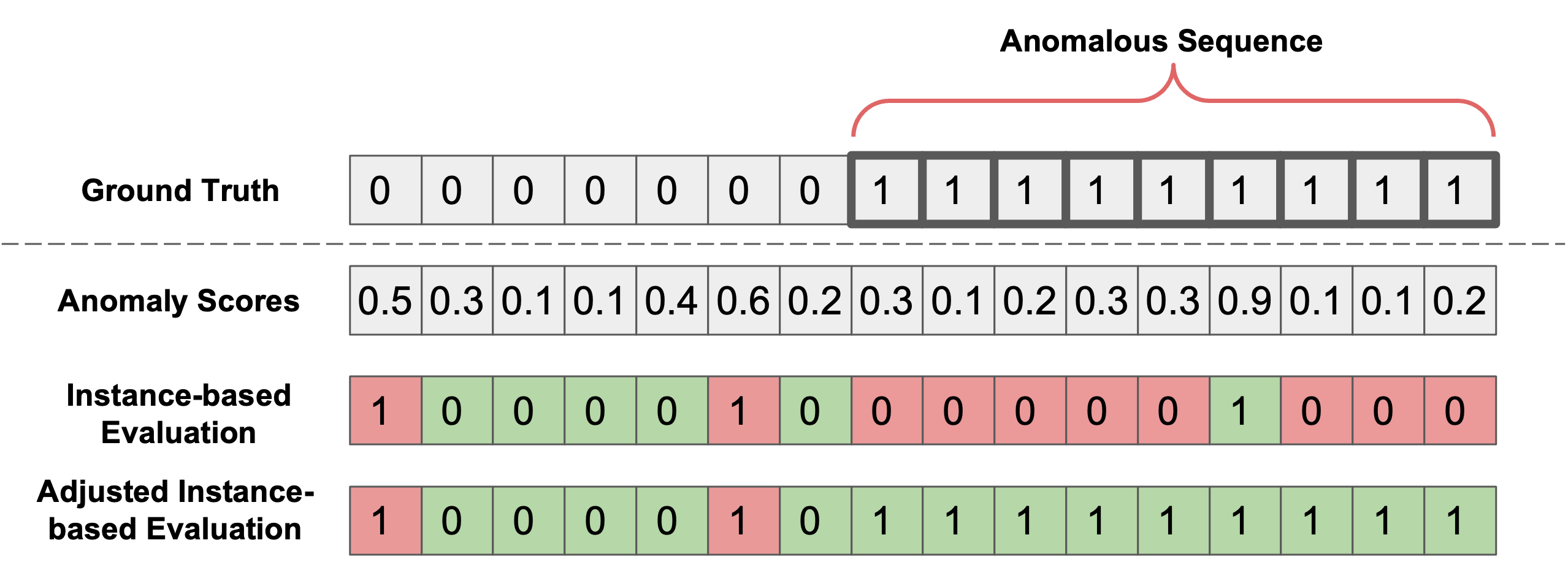}
\caption{The commonly used adjusted instance-based evaluation method. A threshold of $0.5$ is applied to an example sequence of anomaly scores produced by a detection algorithm. The traditional instance-based evaluation compares the anomaly/no-anomaly decision for each instance with the ground truth to determine true/false positive/negative decisions. In the recently proposed and widely used adjusted instance-based evaluation, while errors are penalized once as in the traditional evaluation approach, true detections are greatly amplified by considering all instances in an anomalous sequence as multiple true positives if an alarm is raised during the anomalous sequence. This amplification of true positives causes an artificially inflated F1 score (see Table \ref{tab:faulty}).}
\label{fig:faulty}
\end{figure}

\begin{table*}[!tbh]
    
    \centering
    \begin{small}
    \renewcommand{\multirowsetup}{\centering}
    \setlength{\tabcolsep}{2.15pt}
    \begin{tabular}{c|ccc|ccc|ccc|ccc|ccc}
    \toprule
    Dataset & \multicolumn{3}{c}{SMD} & \multicolumn{3}{c}{MSL} & \multicolumn{3}{c}{SMAP} & \multicolumn{3}{c}{SWaT} & \multicolumn{3}{c}{PSM} \\
    \cmidrule(lr){2-4}\cmidrule(lr){5-7}\cmidrule(lr){8-10}\cmidrule(lr){11-13}\cmidrule(lr){14-16}
    Metric & P & R & F1 & P & R & F1 & P & R & F1 & P & R & F1 & P & R & F1 \\
    \midrule
    OCSVM \cite{scholkopf2001estimating} & {44.34} & {76.72} & {56.19} & {59.78} & {86.87} & {70.82} & {53.85} & {59.07} & {56.34} & {45.39}  & {49.22} & {47.23} & {62.75} & {80.89}  & {70.67}  \\\rule{0pt}{8pt}
    
      IsolationForest \cite{liu2008isolation} & {42.31} & {73.29} & {53.64} & {53.94} & {86.54} & {66.45} & {52.39} & {59.07} & {55.53} & {49.29}  & {44.95} & {47.02} & {76.09} & {92.45}  & {83.48} \\\rule{0pt}{8pt}
          
     LOF \cite{breunig2000lof} & {56.34} & {39.86} & {46.68} & {47.72}  & {85.25} & {61.18} & {58.93} & {56.33} & {57.60} &  {72.15} &{65.43} & {68.62} & {57.89} & {90.49}  & {70.61}  \\\rule{0pt}{8pt}
    
    Deep-SVDD \cite{ruff2018deep} & {78.54} & {79.67} & {79.10} & {91.92} & {76.63} & {83.58}  & {89.93} & {56.02} & {69.04} & {80.42} & {84.45}& {82.39} & {95.41} & {86.49} & {90.73} \\\rule{0pt}{8pt}
     
    DAGMM \cite{zong2018deep} & {67.30} & {49.89} & {57.30} & {89.60} & {63.93} & {74.62} & {86.45} & {56.73} & {68.51} & {89.92}  & {57.84} & {70.40} & {93.49} & {70.03}  & {80.08}  \\\rule{0pt}{8pt}
    
    LSTM-VAE \cite{park2018multimodal} & {75.76} & {90.08}  & {82.30} & {85.49} & {79.94} & {82.62} & {92.20} & {67.75} & {78.10}  & {76.00}  & {89.50} & {82.20} & {73.62} & {89.92} & {80.96} \\\rule{0pt}{8pt}

    BeatGAN \cite{zhou2019beatgan} & {72.90} & {84.09} & {78.10} & {89.75} & {85.42} & {87.53}  & {92.38} & {55.85} & {69.61} & {64.01}  & {87.46} & {73.92} & {90.30} & {93.84} & {92.04} \\\rule{0pt}{8pt}

    OmniAnomaly \cite{li2019enhancing} & {83.68} & {86.82} & {85.22} & {89.02} & {86.37} & {87.67}  & {92.49} & {81.99} & {86.92} & {81.42}  & {84.30} & {82.83} & {88.39} & {74.46} & {80.83} \\\rule{0pt}{8pt}

    InterFusion \cite{li2021multivariate} & {87.02} & {85.43} & {86.22} & {81.28} & {92.70} & {86.62}  & {89.77} & {88.52} & {89.14} & {80.59}  & {85.58} & {83.01} & {83.61} & {83.45} & {83.52} \\\rule{0pt}{8pt}

    THOC \cite{shen2020timeseries} & {79.76} & {90.95} & {84.99} & {88.45} & {90.97} & {89.69} & {92.06} & {89.34} & {90.68} & {83.94}  & {86.36} & {85.13} & {88.14} & {90.99} & {89.54} \\
    
    Anomaly Transformer \cite{xu2021anomaly} & {89.40} & {95.45} & {92.33} & {92.09} & {95.15} & {93.59} & {94.13} & {99.40} & \textbf{{96.69}} & {91.55}  & {96.73} & {94.07} & {96.91} & {98.90} & {97.89} \\
    
    \midrule
    \emph{Random Guess} & {90.54} & {96.43} & \textbf{{93.39}} & {94.95} & {98.13} & \textbf{{96.51}} & {95.43} & {97.74} & {{96.48}} & {93.21}  & {99.47} & \textbf{{96.24}} & {98.83} & {98.14} & \textbf{{98.48}} \\
    \bottomrule
    \end{tabular}
    \end{small}
    \vskip 0.05in
    \caption{Performance of \emph{Random Guess} with probability of alarm $p=0.01$ dominates the state-of-the-art methods in terms of the commonly used adjusted instance-based evaluation, demonstrating its inherent flaw.}\label{tab:faulty}
\end{table*}

Predictive models provide an intuitive way to detect anomalies \cite{space}.
The motivation is that a predictive model trained on nominal data should give statistically similar prediction error for nominal test data, whereas the prediction error is expected to be larger when encountered with anomalous data. Recurrent Neural Networks (RNN) replaced the classical statistical methods such as Autoregressive (AR) and Autoregressive Moving Average (ARMA) models in many applications. 
While the early RNN structures could not utilize long-term dependencies in time series data due to the diminishing gradient problem, Long Short-Term Memory (LSTM) network overcome this problem by introducing a more complex memory unit \cite{LSTM}. Recently, the attention mechanism was proposed to improve the predictive performance of LSTM \cite{attention}. However, more recently, a completely new deep neural network structure called self-attention, also known as transformer, significantly outperformed the LSTM+attention model to become the state-of-the-art predictive model in many applications. While both the attention and self-attention mechanisms are originally proposed for Natural Language Processing (NLP), they were shown to be effective in various other time series data domains \cite{chaudhari2019attentive,clinic}. 

Evaluating performance for detecting sequential anomalies has been traditionally done in the same way as point anomalies using instance-based detection metrics such as AUC and F1 score. However, consecutive anomalous instances are typically caused by the same event, and in real-world applications raising an alarm for such anomalous event is what matters. For instance, after successfully raising alarm for an anomalous event, it is not important in practice to label every anomalous instance. Hence, recent state-of-the-art methods in time series anomaly detection \cite{shen2020timeseries,li2021multivariate,li2019enhancing,ren2019time,su2019robust,xu2018unsupervised,xu2021anomaly} used a different performance evaluation system instead of the traditional instance-based evaluation, which is suitable for point anomalies. In this new evaluation system,
all instances in an anomalous segment are considered as true positives 
if a single alarm is raised in the entire segment, as shown in Fig. \ref{fig:faulty}. While this evaluation system rightfully focuses on the detection of anomalous events/sequences, it fails to provide a meaningful metric for sequential anomalies since it still uses the instance-based F1 score for performance evaluation. In this \emph{adjusted instance-based evaluation}, since errors are penalized once, but detection is rewarded generously, it leads to an inflated F1 metric. 

To illustrate the serious flaw of the adjusted instance-based F1 metric, let us consider using a \emph{Random Guess} method which randomly (e.g., from a uniform distribution) raises an alarm for each instance with probability $p$. Note that this \emph{Random Guess} method makes arbitrary decisions independent of the dataset. The probability of raising a true alarm increases with the duration of anomalous sequence. The exact probabilities are given in Section \ref{sec:problem}. Since its each random detection is amplified by the anomalous sequence duration while its errors are penalized once, such a rudimentary approach is able to achieve very high adjusted instance-based F1 score on the popular benchmark datasets and even outperform the state-of-the-art models, as shown in Table \ref{tab:faulty}. 

Motivated by this limitation, we propose a proper performance metric for sequential anomalies, which also evaluates the timeliness of alarm. Leveraging the high potential of transformer architectures in capturing the long-term dependencies in time series data, we also propose a novel transformer-based time series anomaly detector. 
Our contributions in this paper can be summarized as follows:
\begin{itemize}
    \item A thorough analysis of the inherent flaw in the existing evaluation metric and a proper metric that measures the timeliness and precision of raised alarms.
    \item A novel end-to-end trained transformer algorithm for time series anomaly detection with asymptotic false alarm rate analysis and closed-form expression for detection threshold.  
\end{itemize}

After discussing related works in Section \ref{sec:related} and limitations of the existing evaluation system in Section \ref{sec:problem}, we present the proposed performance metric and the detector in Section \ref{s:Proposed} and the experimental results in Section \ref{s:experiments}. The paper is concluded in Section \ref{s:conclusion}.
\section{Related Work}
\label{sec:related}

The detection of anomalies in time series has been extensively investigated for a long time and remains an active subject due to the great need for more robust methods in complex real-world scenarios \cite{hodge2004survey,chandola2009anomaly}. 
Popular approaches for detecting anomalies in time series data include CNN models \cite{wen2019time,canizo2019multi,munir2018deepant}, RNN models \cite{bontemps2016collective,zhu2017deep,laptev2017time,malhotra2016lstm}, and spectral residuals \cite{microsoft,oprea2021anomaly}. The most recent approaches use the attention-based mechanisms for time series forecasting \cite {liang2018geoman,qin2017dual,zhou2020time}.

Self-attention, which is also known as transformer, is an attention-based neural network architecture without the sequential structure. Current state-of-the results in the NLP domain are obtained by transformer models, e.g., GPT-3 \cite{gpt3}. 
The transformer encoder in \cite{zerveas2021transformer} is used for unsupervised representation learning of multivariate time series data. It outperforms the state-of-the-art time series classification and regression methods. 

The Temporal Fusion Transformer (TFT) presented in \cite{lim2021temporal} is an attention-based model used for multi-horizon forecasting. 
TFT employs recurrent layers for local processing and interpretable self-attention layers for long-term dependency to learn temporal correlations at various scales. In addition, TFT curbs any pointless components by using specialized elements that pick the critical characteristics and a succession of gating layers to get significant performance in various applications.

Based on Generative Adversarial Network (GAN), in \cite{wu2020adversarial}, the authors develop the Adversarial Sparse Transformer, which acts as a generator for learning sparse attention mappings for specific time steps to enhance time series forecasting.  
Using graph learning with the transformer-based network \cite{chen2021learning} proposes learning the graph structure for IoT systems to learn sensor dependencies in multivariate time series datasets automatically.
Informer is another time series forecasting model based on self-attention, which can be used in anomaly detection \cite{zhou2021informer,guo2021data}.


\section{Problem Formulation}
\label{sec:problem}

\begin{figure*}
\centering
\includegraphics[width=.9\textwidth]{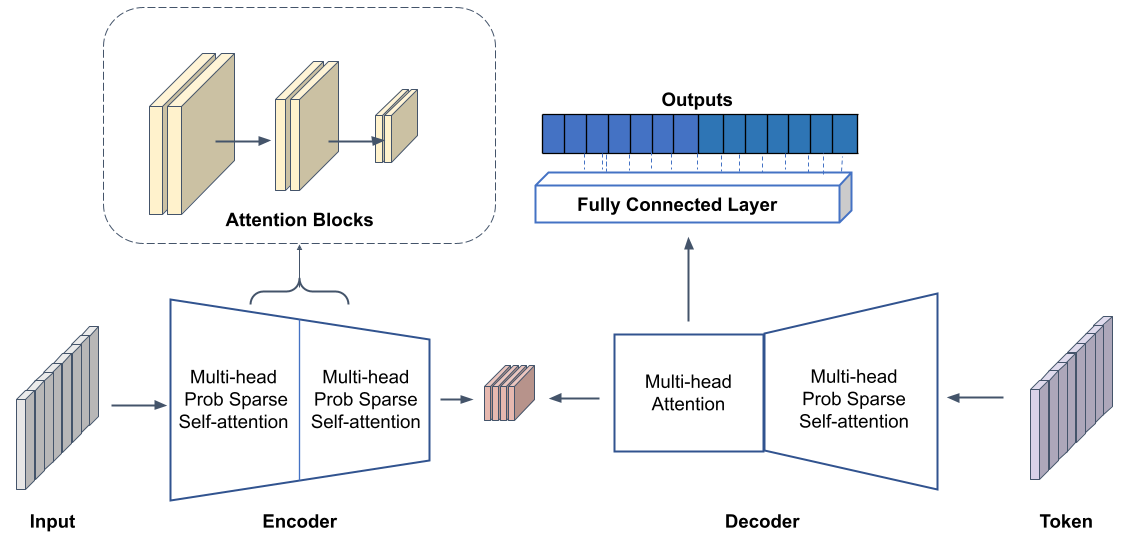}
\caption{Proposed TiSAT architecture}
\label{fig:transformer}
\end{figure*}

\textbf{Sequential Anomaly Detection:} 
Consider a time series $\{X_1,\ldots,X_t,\ldots\}$, which may include anomalous sequences starting and ending at unknown times. Denote the unknown starting time of the $i$th anomalous sequence with $\tau_i$. Since in real-world applications, such as cybersecurity, surveillance, etc. such anomalous sequences are caused by potentially hazardous anomalous events, it is critical to detect such sequences in a timely manner. Controlling the number of false alarms is also crucial to ensure the reliability of the detection system. 
Instead of the traditional instance-based evaluation of performance, which is commonly used for point anomalies and other standard machine learning tasks (e.g., classification, regression), the performance of sequential anomaly detection methods should be evaluated in terms of true/false detection of anomalous sequences. 

\textbf{Flaw of Adjusted Instance-based Evaluation:}
The adjusted instance-based evaluation method has been extensively used in the recent literature (e.g.,  \cite{shen2020timeseries,li2021multivariate,li2019enhancing,ren2019time,su2019robust,xu2018unsupervised,xu2021anomaly}) to compare the state-of-the-art deep learning methods. However, this evaluation method is severely flawed, as illustrated by the high performance of the \emph{Random Guess} algorithm (Table \ref{tab:faulty}). Assuming \emph{Random Guess} raises an alarm with probability $p$, the expected number of false alarms is equal to $Np$, where $N$ is the number of nominal instances. The probability of raising a true alarm within the $i$th anomaly sequence of length $M_i$ is given by $1-Binom(M_i,0,p)$, where $Binom(M_i,0,p)=(1-p)^{M_i}$ is the binomial probability mass function for zero success with $M_i$ trials and $p$ success probability. With the adjusted labeling of the entire sequence as true positive in case of a true alarm, the expected number of true positives is equal to $\sum_{i=1}M_i[1-(1-p)^{M_i}]$. 
Consequently, the expected number of false negatives is given by $\sum_{i=1}M_i(1-p)^{M_i}$. Hence, the expected precision and recall are
\begin{align}
\begin{split}
    \text{Precision}_{\text{RandomGuess}} &= \frac{\sum_{i=1}M_i[1-(1-p)^{M_i}]}{\sum_{i=1}M_i[1-(1-p)^{M_i}]+Np} \\
    \text{Recall}_{\text{RandomGuess}} &= \frac{\sum_{i=1}M_i[1-(1-p)^{M_i}]}{\sum_{i=1}M_i}.
\end{split}
\end{align}
As the duration $M_i$ of anomalous sequences increase, the expected number of true positives increases significantly, making both precision and recall approach to $1$. Note the insignificant effect of false positives ($Np$) since they are penalized once. In the popular benchmark datasets, there are long anomalous sequences lasting thousands of instances, and thus we observe the high precision, recall, and F1 scores in Table \ref{tab:faulty} with $p=0.01$.

\section{Proposed Approach}
\label{s:Proposed}

In this section, we first present a proper performance metric for sequential anomalies and then our transformer based anomaly detection approach. 

\subsection{Performance Evaluation}


\textbf{Sequence Detection Delay:}
Given $\tau_i$ as the starting time of an anomalous event $i$ and $T_i\ge\tau_i$ as the alarm time, we can empirically formulate the average detection delay as 
\begin{equation}
\label{eq:add}
    \text{ADD}=\frac{1}{S}\sum_{i=1}^S (T_i-\tau_i),
\end{equation}
where $S$ denotes the number of anomalous events. Since most anomalies indicate critical incidents, it might be essential to detect an anomalous event within a certain time period. Hence, if no alarm is raised within the duration $[\tau_i,\tau_i+\delta_{\text{max}}]$ after anomalous activity $i$ happens, we set the delay to the maximum tolerable delay $\delta_{\text{max}}$. Here, it is important to note that minimizing the detection delay is analogous to the the more commonly used objective of maximizing the true positive rate, except it assigns a more specific cost of detection delay $\delta_i=T_i-\tau_i$.  

\textbf{Sequence Alarm Precision:}
Our second objective emphasizes on maximizing the number of anomalous events being detected with respect to the total number of alarms, similar to the well-known precision metric. However, in contrast to the instance-based precision metric, our metric focuses on the detection of true anomalous \emph{sequences}, and hence only focuses on detecting the anomalous event onset accurately. If an alarm is raised before an alarm even begins, i.e.,  $T^j\le \tau_i$, then it is considered as a false alarm. 

Empirically, the alarm precision is computed as 
\begin{equation}
\label{eq:prec}
    P=\frac{1}{\hat{S}}\sum_{j=1}^{\hat{S}} \ind ={T_j \in \cup [\tau_i,\tau_i+\delta_{\text{max}}]},
\end{equation}
where $\ind{\cdot}$ denotes the indicator function, $\hat{S}=|\{T_j\}|$ is the number of all alarms, and $|\cdot|$ denotes the cardinality of a set. 

\textbf{Sequence Precision Delay:} Finally, we present a new metric called \emph{Sequence Precision Delay} that combines the sequence based average detection delay with sequence alarm precision in order to achieve a single metric for easily comparing time series algorithms. The SPD statistic quantifies the area under the Precision vs. normalised ADD (NADD) curve, much like the common AUC metric does for TPR and FPR. To map ADD into $[0,1]$, we normalize it by the maximum delay, i.e., $\text{NADD}=\text{ADD}/\delta_{\text{max}}$. Mathematically, SPD is given by
\begin{equation}
\label{eq:met}
    \text{SPD} = \int_{0}^{1} P(\alpha) ~\text{d}\alpha,
\end{equation}
where $\alpha$ denotes NADD, and $P$ denotes the precision. A highly successful algorithm with an SPD value close to $1$ must have high precision and low delay in its alarms.

Most existing approaches leverage an RNN based model for time series forecasting, and compute the \emph{residuals}, i.e. the prediction or reconstruction error to determine if an observation is anomalous or not. However, it is shown in \cite{zhou2021informer} that RNN based approaches suffer in long sequence time series forecasting. To this end, we propose a novel transformer based approach called \emph{Time Series Anomaly Transformer (TiSAT)}, which is superior in capturing long range temporal dependencies. We next discuss our proposed approach in detail. 

\subsection{Time Series Anomaly Transformer (TiSAT)}

The overall structure of TiSAT is shown in Fig. \ref{fig:transformer}. The proposed approach utilizes the the ProbSparse mechanism discussed in the Informer architecture \cite{zhou2021informer} as compared to the self-attention mechanism proposed in the Vanilla Transformer \cite{vaswani2017attention} for reducing computational complexity. Traditionally, the self attention mechanism for a $d$-dimensional input is defined as 

\begin{equation}
    A(Q, K, V) = Softmax\left(\frac{QK^T}{\sqrt{d}}\right)V
\end{equation}
where $Q$, $K$ and $V$ represent the query, key and value respectively.  
However, it was recently observed that only a few key and value pairs contribute to the attention score, rendering a majority of the computed dot products as worthless. Hence, we propose using a probabilistic attention mechanism, since it reduces the computational complexity from $\mathcal{O}(L^2)$ to $\mathcal{O}(LlogL)$. Particularly, since only a subset of the query/value tensors require costly operations, ProbSparse attention allows each key to focus on the most important queries rather than all of them. The ProbSparse attention is given by
\begin{equation}
    A(Q, K, V) = Softmax\left(\frac{\bar QK^T}{\sqrt{d}}\right)V
\end{equation}
where $\bar Q$ is a sparse matrix consisting of the top queries. As shown in \cite{zhou2021informer}, the measurement of sparsity between a query $q_i$ and all keys $K$ is given by

\begin{equation}
    M(q_i,K) = \log\left(\sum_{j=1}^{L_K} e^\frac{q_ik_j^T}{\sqrt{d}}\right) - \frac{1}{L_K}\sum_{j=1}^{L_K} \frac{q_ik_j^T}{\sqrt{d}}
\end{equation}

\textbf{Encoder:} The input to the proposed architecture at a time instance $t$ is a matrix representation of the $d$-dimensional time series over a sequence of $L$ instances given by $\mathcal{Z}^t$. Following the dilated convolution approach proposed in \cite{yu2017dilated,gupta2017dilated}, we employ a distilling procedure to extract a focused self-attention feature map. The distilling function from the $j^{th}$ layer towards the $(j+1)^{th}$ layer is given by
\begin{equation}
    \mathcal{Z}_{j+1}^t = \text{MaxPool ELU}(Conv1d([Z_j])) 
\end{equation}
where $\left [ \cdot  \right ]$ represents the attention block, which is followed by a 1-D convolution filter with a kernel width of 3. The number of self-attention blocks are progressively decreased and then finally concatenated to form the final representation of the encoder.  

\textbf{Decoder:} We leverage the canonical decoder structure proposed by \cite{vaswani2017attention} and consists of two similar self-attention modules. The input to the decoder architecture is given by concatenating the start token ($Z_{token})$ and a placeholder for the target sequence ($Z_{target}$) as follows  
\begin{equation}
  Z^t_{de} = Concat(Z_{token},Z_{target})
\end{equation}
This is then passed to a dense fully connected layer. The network is trained using the mean squared error loss by propagating it through the decoder and encoder.

\subsection{Anomaly Detection Framework}

We propose an online and non-parametric detection approach for detecting persistent and abrupt anomalies using the transformer output. Due to the sequential (persistent) nature of time series anomalies, we need an approach which accumulates the evidence over time and then makes a decision instead of a hard threshold on the anomaly score for each instant. 
To this end, we propose using a nonparametric sequential algorithm based on $k$ nearest neighbors ($k$NN). First, the algorithm trains on a set of nominal historic observations in an offline fashion and then tests the incoming observations until it detects a change in the observations with respect to the nominal baseline. In the training phase, assuming a training set $\cX_N$ consisting of $N$ nominal data instances, it randomly partitions $\cX_N$ into two sets $\cX_{N_1}$ and $\cX_{N_2}$, where $N_1+N_2=N$. Then, for each point in $\cX_{N_1}$ it finds the Euclidean distance to each point in $\cX_{N_2}$ as $\{d_1,d_2,\dots,d_{N_1}\}$. The (1-$\alpha$)th percentile $d_\alpha$ is used as a baseline statistic during the testing phase, where $\alpha$ is the statistical significance level (e.g., $0.05$). During testing, we compute the anomaly evidence for each instance as
\begin{equation}
\label{e:Dt}
    D_t = d_t^m - d_\alpha^m ,
\end{equation}
where $d_t$ is the $k$NN distance between the test point at time instance $t$ and $\cX_{N_2}$ and $m$ is the dimensionality of the transformer output. 

The anomaly evidences are accumulated over time
\begin{equation}
\label{e:stk}
s_t = \max\{s_t + D_t, 0\}, s_0 = 0.
\end{equation}
and an alarm is raised when the anomaly statistic $s_t$ exceeds a threshold $h$, i.e., at time
\begin{equation}
\label{e:T}
T = \min\{t : s_t^k \leq h \}.
\end{equation}

Here, $T$ is the minimum time required for the accumulated evidence $s_t^k$ to be sufficiently high to raise an alarm based on a detection threshold $h$. The detection threshold $h$ manifests a trade-off between minimizing the detection delay and minimizing the false alarm.

For an anomaly detection algorithm to be implemented in a practical setting, a clear procedure is necessary for selecting the decision threshold such that it satisfies a desired false alarm rate. The reliability of an algorithm in terms of false alarm rate is crucial for minimizing human involvement. To provide such a performance guarantee for the false alarm rate, we derive an asymptotic upper bound on the average false alarm rate of the proposed algorithm.

\begin{thm}
\label{thm:1}
The false alarm rate of the proposed algorithm is asymptotically (as $N_2 \rightarrow \infty$) upper bounded by 
\begin{equation}
    FAR \leq e^{-\omega_0h}, 
\end{equation}
where $h$ is the decision threshold, and $\omega_0>0$ is given by
\begin{align}
    \label{e:thm}
    \omega_0 &= v_m - \theta -\frac{1}{\phi} \mathcal{W}\left( -\phi \theta e^{-\phi\theta } \right), \\
    \theta &= \frac{v_m}{e^{v_m d_\alpha^m}}.\nonumber
\end{align}
In \eqref{e:thm}, $\mathcal{W}(\cdot)$ is the Lambert-W function, $v_m=\frac{\pi^{m/2}}{\Gamma(m/2+1)}$ is the constant for the $m$-dimensional Lebesgue measure (i.e., $v_m d_\alpha^m$ is the $m$-dimensional volume of the hyperball with radius $d_\alpha$), and $\phi$ is the upper bound for $D_t$.
\end{thm}
\textit{Proof.} In \cite{basseville1993detection}[page 177], for CUSUM-like algorithms with independent increments, such as TiSAT with independent $D_t$, a lower bound on the average false alarm period is given as follows 
\[
E_\infty[T] \geq e^{\omega_0h},
\]
where $h$ is the detection threshold, and $\omega_0 \geq 0$ is the solution to $E[e^{\omega_0D_t}] = 1$. 

To analyze the false alarm period, we need to consider the nominal case. 
The anomaly evidence in the nominal case does not necessarily depend on the previous selections due to lack of an anomaly which could correlate the evidences. Hence, in the nominal case, it is safe to assume that $D_t$ is independent over time.

We firstly derive the asymptotic distribution of the instance-level anomaly evidence $D_t$ in the absence of anomalies. Its cumulative distribution function is given by 
\[
P(D_t \leq y) = P(d_t^m \leq d_\alpha^m + y).
\]
It is sufficient to find the probability distribution of $d_t^m$, the $m$th power of the $k$NN distance at time $t$. As discussed above, we have independent $m$-dimensional vectors $\{X_t\}$ over time, which form a Poisson point process. The nearest neighbor ($k=1$) distribution for a Poisson point process is given by
\[
P(d_t\} \leq r) = 1 - \exp(-\Lambda(b(X_t,r)))
\]
where $\Lambda(b(X_t,r))$ is the arrival intensity (i.e., Poisson rate measure) in the $m$-dimensional hypersphere $b(X_t,r)$ centered at $X_t$ with radius $r$ \cite{chiu2013stochastic}. Asymptotically, for a large number of training instances as $N_2\to\infty$, under the null (nominal) hypothesis, $d_t$ of $X_t$ takes small values, defining an infinitesimal hyperball with homogeneous intensity $\lambda=1$ around $X_t$. Since for a homogeneous Poisson process the intensity is written as $\Lambda(b(X_t,r)) = \lambda |b(X_t,r)|$ \cite{chiu2013stochastic}, where $|b(X_t,r)| = \frac{\pi^{m/2}}{\Gamma(m/2+1)}r^m = v_m r^m$ is the Lebesgue measure (i.e., $m$-dimensional volume) of the hyperball $b(X_t,r)$, we rewrite the nearest neighbor distribution as
\begin{align*}
P(d_t \le r) &= 1-\exp\left( -v_m r^m \right),
\end{align*}
where $v_m = \frac{\pi^{m/2}}{\Gamma(m/2+1)}$ is the constant for the $m$-dimensional Lebesgue measure. 
Now, applying a change of variables we can write the probability density of $d_t^m$ and $D_t$ as
\begin{align}
f_{d_t^m}(y) &= \frac{\partial}{\partial y} \left[1-\exp\left( -v_m y \right) \right], \\
&= v_m \exp(-v_m y), \\
\label{eq:pdf}
f_{D_t}(y) &= v_m \exp(-v_m d_{\alpha}^m) \exp(-v_m y)
\end{align}

Using the probability density derived in \eqref{eq:pdf}, $E[e^{\omega_0D_t}] = 1$ can be written as
\begin{align}
1 &= \int_{-d_\alpha^m}^{\phi} e^{\omega_0 y}v_me^{-v_m d_\alpha^m}e^{-v_my}dy, \\
\frac{e^{v_md_\alpha^m}}{v_m} &= \int_{-d_\alpha^m}^{\phi} e^{(\omega_0-v_m)y}dy, \\
&= \frac{e^{(\omega_0-v_m)y}}{\omega_0-v_m}\Biggr|_{-d_\alpha^m}^{\phi}, \\ 
&=\frac{e^{(\omega_0-v_m)\phi} - e^{(\omega_0-v_m)(-d_\alpha^m)}}{\omega_0-v_m},
\end{align}
where $-d_\alpha^m$ and $\phi$ are the lower and upper bounds for $D_t=d_t^m-d_\alpha^m$. The upper bound $\phi$ is obtained from the training set. 

As $N_2\to\infty$, since the $m$th power of $(1-\alpha)$th percentile of nearest neighbor distances in training set goes to zero, i.e., $d_\alpha^m \to 0$, we have 
\begin{align}
e^{(\omega_0-v_m)\phi} &= \frac{e^{v_m d_\alpha^m}}{v_m}(\omega_0-v_m) + 1. 
\end{align}

We next rearrange the terms to obtain the form of $e^{\phi x} = a_0(x+\theta)$ where $x=\omega_0-v_m$, $a_0=\frac{e^{v_m d_\alpha^m}}{v_m}$, and $\theta=\frac{v_m}{e^{v_m d_\alpha^m}}$. The solution for $x$ is given by the Lambert-W function \cite{scott2014asymptotic} as $x = -\theta - \frac{1}{\phi} \mathcal{W}(-\phi e^{-\phi\theta}/a_0)$, hence
\begin{align}
\omega_0 = v_m - \theta -\frac{1}{\phi} \mathcal{W}\left( -\phi \theta e^{-\phi\theta } \right). 
\end{align}

Finally, since the false alarm rate (i.e., frequency) is the inverse of false alarm period $E_\infty[T]$, we have 
\[
FAR \leq e^{-\omega_0 h},
\]
where $h$ is the detection threshold, and $\omega_0$ is given above.

\begin{figure*}[tbh]
\minipage{0.32\textwidth}
  \includegraphics[width=\linewidth,height=4.5cm]{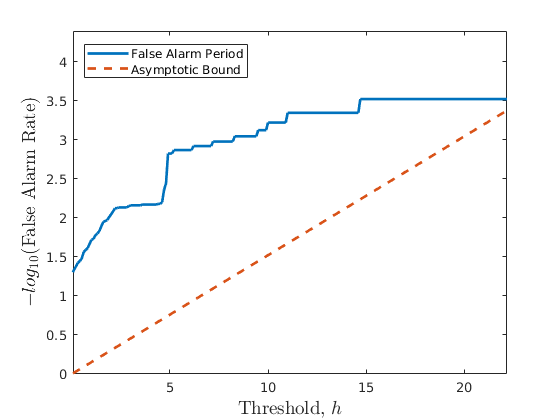}

\endminipage\hfill
\minipage{0.32\textwidth}
  \includegraphics[width=\linewidth,height=4.5cm]{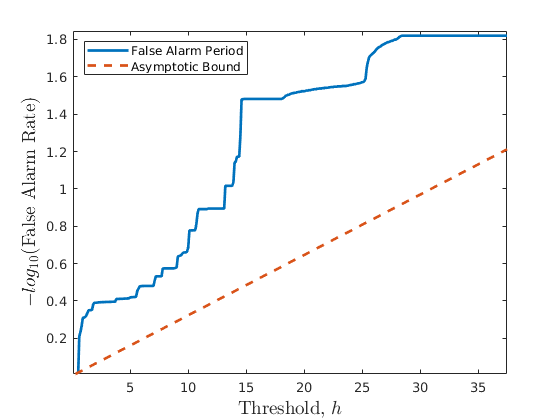}

\endminipage\hfill
\minipage{0.32\textwidth}%
  \includegraphics[width=\linewidth,height=4.5cm]{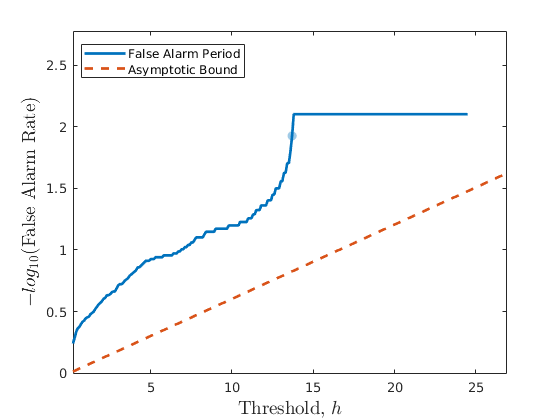}

\endminipage
\caption{Actual false alarm periods vs. derived lower bounds for the PSM, SMAP and SWAT datasets respectively.}
\label{f:threshold}
\end{figure*}

Specifically, $v_m$ is directly computed using the dimensionality $m$, $d_\alpha$ comes from the training phase, $\phi$ is also found in training, and finally there is a built-in Lambert-W function in popular programming languages such as Python and Matlab.
Hence, given the training data, $\omega_0$ can be easily computed, and based on Theorem \ref{thm:1}, the threshold $h$ can be chosen to asymptotically achieve the desired false alarm period as follows
\begin{equation}
h = \frac{-\log (FAR)}{\omega_0}
\end{equation}

We finally present the comparison between the bound for false alarm rate derived in Theorem \ref{thm:1} and the empirical false alarm period in Fig. \ref{f:threshold}. The figure depicts the logarithm of false alarm period, which is the inverse of false alarm rate, for clarity. Hence, the upper bound on the false alarm rate becomes the lower bound on the false alarm period in this scenario.

\section{Experiments}
\label{s:experiments}

\textbf{Datasets:} Most of the existing works evaluate their performance on the following datasets:

\begin{itemize}
    \item \textbf{SMD (Server Machine Dataset)}: The SMD dataset is collected from a large internet company and consists of data collected over 5 weeks with 38 dimensions.
    \item \textbf{PSM (Pooled Server Metrics)}: The PSM dataset is proposed by eBay and consists of 26 dimensional data captured internally from application server nodes.
    \item \textbf{MSL} (Mars Science Laboratory rover) and \textbf{SMAP} (Soil Moisture Active Passive satellite): The MSL and SMAP datasets are provided by NASA and consists of telemetry data and anomalies featuring 55 and 25 dimensions respectively. Since most of the dimensions are categorical, we only focus on the telemetry values.   
    \item \textbf{SWaT}: The SWaT dataset is collected in an industrial setting and features data collected from a sewage water treatment facility. The dataset is collected over an entire week and consists of 51 dimensions, where the anomalies are caused due to cyberattacks.    
\end{itemize}
Each dataset includes training, validation and testing subsets. Anomalies are only labeled in the testing subset.

\textbf{Implementation Details:} Following the sliding window approach used in existing works, the input to the proposed TiSAT model is a sub-series with a window size of 100. The TiSAT model encoder consists of a 3-layer stack followed by a 1-layer stack and the decoder consists of a 2-layer stack. We train the model using Adam optimizer, and the learning rate is set as 1$e^{-4}$. We train the model for 4 epochs with a batch size of 64. We normalize all the datasets between [0,1] using minmax normalization.

\textbf{Results:} We extensively evaluate the performance of the proposed approach on the five publicly available benchmark datasets using the proposed SPD metric. We compare our approaches with classical algorithms such as OC-SVM, IsolationForest and LOF, as well as time series forecasting based approaches such as ARIMA and LSTM. As shown in Table \ref{tab:main_results2}, the proposed TiSAT model significantly outperforms all other approaches. While there have been several neural network based approaches \cite{shen2020timeseries,li2021multivariate,li2019enhancing,ren2019time,su2019robust,xu2018unsupervised,xu2021anomaly} proposed recently, all of them present their results using the faulty adjusted instance-based metric shown in Fig. \ref{fig:faulty} (See Table \ref{tab:faulty} for the inherent flaw of this metric). Since most of them do not have their codes available, it is not feasible for us to compare with them.

\begin{table}[!tbh]
    
    \centering
    \begin{small}
    \renewcommand{\multirowsetup}{\centering}
    \begin{tabular}{c|c|c|c|c|c}
    \toprule
    Dataset & \multicolumn{1}{c}{SMD} & \multicolumn{1}{c}{MSL} & \multicolumn{1}{c}{SMAP} & \multicolumn{1}{c}{SWaT} & \multicolumn{1}{c}{PSM} \\
    Metric (SPD)  \\
    \midrule
    OCSVM \cite{scholkopf2001estimating} & {21.08} & {20.15} & {12.82} & {17.21} & {19.44}  \\
    
      IsolationForest \cite{liu2008isolation} & {19.38} & {16.50} & {9.26} & {12.83} & {18.98} \\
          
     LOF \cite{breunig2000lof} & {17.25} & {14.29} & {18.21} & {13.68}  & {20.14} \\
     
     ARIMA & {27.64} & {24.13} & {22.87} & {29.35}  & {26.47} \\
     
     LSTM \cite{malhotra2016lstm} & {28.31} & {23.85} & {27.54} & {32.97}  & {28.41} \\
    \midrule
    Ours & \textbf{{52.70}} & \textbf{{33.40}} & \textbf{{29.35}} & \textbf{{46.23}} & \textbf{{38.58}}  \\
    \bottomrule
    \end{tabular}
    \end{small}
    \vskip 0.05in
    \caption{Comparison with existing approaches using the proposed SPD metric.} \label{tab:main_results2}
\end{table}
\section{Conclusion}
\label{s:conclusion}

In this work we identified a crucial shortcoming in the existing evaluation criterion used by most recent approaches for time series anomaly detection. To rectify the evaluation method, we presented a novel performance metric which measures the timeliness and precision of detection methods. Moreover, we proposed a novel transformer based approach called TiSAT for unsupervised time series anomaly detection and provided an asymptotic false alarm rate analysis for TiSAT. This analysis leads to a closed-form expression for the detection threshold, which was emppirically corroborated on benchmark datasets.  
We comprehensively evaluated the proposed approach and showed that TiSAT is able to achieve state-of-the-art performance on benchmark datasets.

\bibliographystyle{ieeetr}
\bibliography{egbib}

\end{document}